\documentclass[10pt,twocolumn,letterpaper]{article}

\usepackage{iccv}
\usepackage{times}
\usepackage{epsfig}
\usepackage{graphicx}
\usepackage{amsmath}
\usepackage{amssymb}
\usepackage{epsfig}
\usepackage[accsupp]{axessibility}
\usepackage{float}
\usepackage{subfig}
\usepackage{booktabs}
\usepackage{times}
\usepackage{url}
\usepackage{amsthm}
\usepackage{algorithm}
\usepackage{algorithmic}
\usepackage{multirow}
\usepackage{amsfonts} 
\usepackage{bbding}
\usepackage{makecell}
\usepackage{colortbl}
\usepackage{xcolor}
\usepackage{microtype}
\usepackage{bm}
\usepackage{wrapfig}
\usepackage{diagbox}

\usepackage{pifont}

\definecolor{cred}{HTML}{FF6B6B}
\definecolor{cyellow}{HTML}{FEC260}
\definecolor{cgreen}{HTML}{70AD47}
\definecolor{cblue}{HTML}{4D96FF}
\definecolor{cpurple}{HTML}{2A0944}
\definecolor{ggray}{RGB}{127,127,127}
\definecolor{aliceblue}{rgb}{0.94, 0.97, 1.0}

\usepackage[breaklinks=true,bookmarks=false,colorlinks]{hyperref}

\iccvfinalcopy 

\def\iccvPaperID{****} 

\ificcvfinal\fi

\makeatletter
\newcommand{\ssymbol}[1]{$^{\@fnsymbol{#1}}$}
\makeatother

\begin{document}

\title{DiffusionRet: Generative Text-Video Retrieval with Diffusion Model}

\author{
    Peng Jin$^{1,3}$\footnotemark[1] \ \
    Hao Li$^{1,3}$\footnotemark[1] \ \ 
    Zesen Cheng$^{1,3}$ \ 
    Kehan Li$^{1,3}$ \ 
    Xiangyang Ji$^{4}$ \ \and 
    Chang Liu$^{4}$ \ 
    Li Yuan$^{1,2,3}$\footnotemark[2] \ \
    Jie Chen$^{1,2,3}$\footnotemark[2] \\[3pt]
    \small{$^1$School of Electronic and Computer Engineering, Peking University, Shenzhen, China} \quad
    \small{$^2$Peng Cheng Laboratory, Shenzhen, China} \\
    \small{$^3$AI for Science (AI4S)-Preferred Program, Peking University Shenzhen Graduate School, Shenzhen, China} \\
    \small{$^4$Department of Automation and BNRist, Tsinghua University, Beijing, China} \\
    \footnotesize{\{jp21, cyanlaser, kehanli\}@stu.pku.edu.cn} \quad \footnotesize{\{lihao1984, yuanli-ece\}@pku.edu.cn} \quad
    \footnotesize{\{liuchang2022, xyji\}@tsinghua.edu.cn \quad chenj@pcl.ac.cn}
}

\maketitle
\ificcvfinal\thispagestyle{empty}\fi

\newcommand{\myparagraph}[1]{\textbf{#1}\hspace{1.8ex}}
\newcommand{\mysubparagraph}[1]{\textit{#1}\hspace{1.8ex}}
\renewcommand{\thefootnote}{\fnsymbol{footnote}}

\footnotetext[1]{Equal contribution.}
\footnotetext[2]{Corresponding author: Li Yuan, Jie Chen.}

\begin{abstract}
Existing text-video retrieval solutions are, in essence, discriminant models focused on maximizing the conditional likelihood, i.e., $p(\textit{candidates}|\textit{query})$. While straightforward, this de facto paradigm overlooks the underlying data distribution $p(\textit{query})$, which makes it challenging {to identify out-of-distribution data}. To address this limitation, we creatively tackle this task from a generative viewpoint and model the correlation between the text and the video as their joint probability $p(\textit{candidates},\textit{query})$. This is accomplished through a \underline{diffusion}-based text-video \underline{ret}rieval framework~(DiffusionRet), which models the retrieval task as a process of gradually generating joint distribution from noise. During training, DiffusionRet is optimized from both the generation and discrimination perspectives, with the generator being optimized by generation loss and the feature extractor trained with contrastive loss. In this way, DiffusionRet cleverly leverages the strengths of both generative and discriminative methods. Extensive experiments on five commonly used text-video retrieval benchmarks, including MSRVTT, LSMDC, MSVD, ActivityNet Captions, and DiDeMo, with superior performances, justify the efficacy of our method. More encouragingly, without any modification, DiffusionRet even performs well in out-domain retrieval settings. We believe this work brings fundamental insights into the related fields. Code is available at \href{https://github.com/jpthu17/DiffusionRet}{https://github.com/jpthu17/DiffusionRet}.
\end{abstract}

\begin{figure}[tbp]
\centering
\includegraphics[width=1.\linewidth]{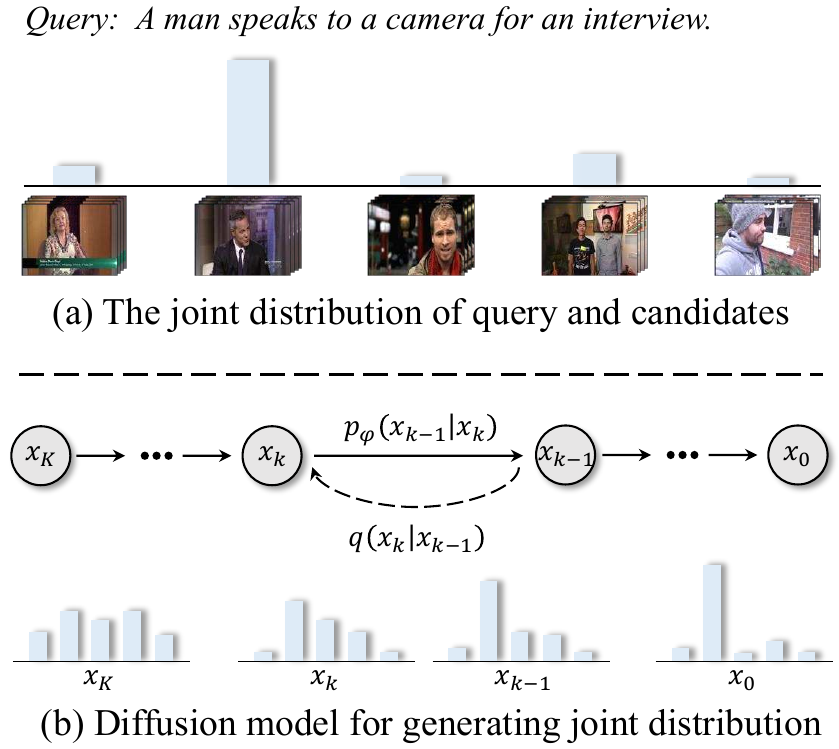}
\caption{\textbf{Diffusion model for text-video retrieval.} (a)~We propose to model the correlation between the query and the candidates as their joint probability. (b) The diffusion model has demonstrated remarkable generative power in various fields, and due to its coarse-to-fine nature, we utilize the diffusion model for joint probability generation.}
\label{fig1}
\end{figure}

\section{Introduction}
In recent years, text-video retrieval has made significant progress, allowing humans to associate textual concepts with video entities and vice versa~\cite{wang2021t2vlad,ijcai/WangZCCZPGWS21}. Existing methods for video-text retrieval typically model the cross-modal interaction as discriminant models~\cite{gabeur2020multi,luo2021clip4clip}. Under the discriminant paradigm based on contrastive learning~\cite{oord2018representation}, the primary focus of mainstream methods is to improve the dense feature extractor to learn better representation. This has led to the emergence of a large number of discriminative solutions~\cite{jin2022expectation,bain2021frozen,liu2022ts2,fang2021clip2video}, and recent advances in large-scale vision-language pre-training models~\cite{radford2021learning,li2022blip} have pushed their state-of-the-art performance even further.

However, from a probabilistic perspective, discriminant models only learn the conditional probability distribution, \ie, $p(\textit{candidates}|\textit{query})$. This leads to a limitation of discriminant models that they fail to model the underlying data distribution~\cite{bernardo2007generative,liang2022gmmseg}, and their latent space contains fewer intrinsic data characteristics $p(\textit{query})$, making it difficult to achieve good generalization on unseen data~\cite{liang2022gmmseg}. In contrast to discriminant models, generative models capture the joint probability distribution of the query and candidates, \ie, $p(\textit{candidates},\textit{query})$, which allows them to project data into the correct latent space based on the semantic information of the data. As a typical consequence, generative models are often more generalizable and transferrable than discriminant models. Recently, generative models have made significant progress in various fields, such as generating high-quality synthetic images~\cite{dhariwal2021diffusion,brock2018large,razavi2019generating}, natural language~\cite{brown2020language}, speech~\cite{oord2016wavenet}, and music~\cite{dhariwal2020jukebox}. One of the most popular generative paradigms is the diffusion model~\cite{sohl2015deep,ho2020denoising,dhariwal2021diffusion}, which is a type of likelihood-based model that gradually remove noise via a learned denoising model to generate the signal. In particular, the coarse-to-fine nature of the diffusion model enables it to progressively uncover the correlation between text and video, making it a promising solution for cross-modal retrieval. Therefore, we argue that it is time to rethink the current discriminant retrieval regime from a generative perspective, utilizing the diffusion model.

To this end, we propose a novel \underline{diffusion}-based text-video \underline{ret}rieval framework, called DiffusionRet, which addresses the limitations of current discriminative solutions from a generative perspective. As shown in Fig.~\ref{fig1}, we model the retrieval task as a process of gradually generating joint distribution from noise. Given a query and a gallery of candidates, we adopt the diffusion model to generate the joint probability distribution $p(\textit{candidates},\textit{query})$. To improve the performance of the generative model, we optimize the proposed method from both generation and discrimination perspectives. During training, the generator is optimized by common generation loss, \ie, Kullback-Leibler divergence~\cite{kullback1997information}. Simultaneously, the feature extractor is trained with the contrastive loss, \ie, InfoNCE loss~\cite{oord2018representation}, to enable discriminative representation learning. In this way, DiffusionRet has both the high performance of discriminant methods and the generalization ability of generative methods.

The proposed DiffusionRet has two compelling advantages: \textbf{First}, the generative paradigm of our method makes it inherently generalizable and transferrable, enabling DiffusionRet to adapt to out-of-domain samples without requiring additional design. \textbf{Second}, the iterative refinement property of the diffusion model allows DiffusionRet to progressively enhance the retrieval results from coarse to fine. Experimental results on five benchmark datasets for text-video retrieval, including MSRVTT~\cite{xu2016msr}, LSMDC~\cite{rohrbach2015a}, MSVD~\cite{chen2011collecting}, ActivityNet Captions~\cite{krishna2017dense}, and DiDeMo~\cite{anne2017localizing}, demonstrate the advantages of DiffusionRet. To further evaluate the generalization of our method to unseen data, we propose a new out-domain retrieval task. In the out-domain retrieval task~\cite{chen2020fine}, labeled visual data with paired text descriptions are available in one domain (the ``source''), but no data are available in the domain of interest (the ``target''). Our method not only represents a novel effort to promote generative methods for in-domain retrieval, but also provides evidence of the merits of generative approaches in challenging out-domain retrieval settings. The main contributions are as follows:

\begin{itemize}
    \item To the best of our knowledge, we are the first to tackle the text-video retrieval from a generative viewpoint. Moreover, we are the first to adapt the diffusion model for cross-modal retrieval. 
    
    \item Our method achieves new state-of-the-art performance on text-video retrieval benchmarks of \textit{MSRVTT}, \textit{LSMDC}, \textit{MSVD}, \textit{ActivityNet Captions} and \textit{DiDeMo}.
    
    \item More impressively, our method performs well on out-domain retrieval without any modification, which may have a significant impact on the community.
\end{itemize}

\section{Related Work}
\myparagraph{Text-Video Retrieval.} 
Text-video retrieval is one of the most popular cross-modal tasks~\cite{li2022neighborhood,li2022deep,li2023weakly,cheng2023wico,li2023tg}. Most existing text-video retrieval works~\cite{fang2023uatvr,wang2022disentangled,li2023momentdiff,li2022dual} employ a mapping technique that places both text and video inputs in the same latent space for direct similarity calculation. For instance, CLIP4Clip~\cite{luo2021clip4clip} transfers knowledge from the text-image pre-training model, \ie, CLIP~\cite{radford2021learning}, to enhance video representation. EMCL-Net~\cite{jin2022expectation} improves representation capabilities by bridging the gap between text and video. HBI~\cite{jin2022video} models video-text as game players with multivariate cooperative game theory to handle the uncertainty during fine-grained semantic interaction. These methods are discriminant models and focus primarily on maximizing the conditional likelihood $p(\textit{candidates}|\textit{query})$, without considering the underlying data distribution $p(\textit{query})$. As a result, identifying out-of-distribution data becomes a challenging task. In contrast, our DiffusionRet adopts a generative viewpoint to tackle the task and provides evidence of the merits of generative approaches in a challenging, out-domain retrieval. To the best of our knowledge, we are the first to tackle text-video retrieval from a generative viewpoint.

\begin{figure*}[tbp]
\centering
\includegraphics[width=1.\linewidth]{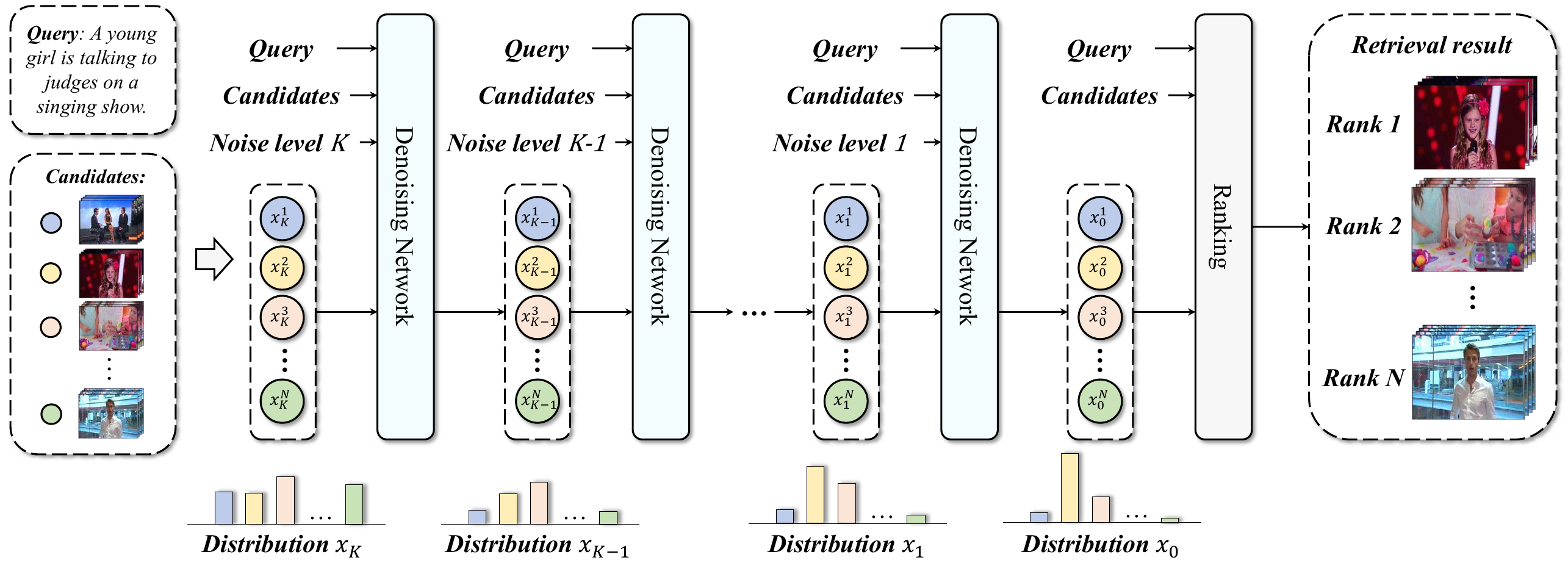}
\vspace{-1.5em}
\caption{\textbf{Our DiffusionRet framework for generative text-video retrieval.} We model the retrieval task as a process of gradually generating joint distribution from Gaussian noise. In contrast to the prior works, which typically optimize the posterior probabilities $p(v|t)+p(t|v)$, our method builds the joint probabilities $p(v,t)$.}
\label{fig2}
\end{figure*}

\myparagraph{Diffusion models.} Diffusion models~\cite{sohl2015deep,ho2020denoising,dhariwal2021diffusion} are a type of neural generative model that uses the stochastic diffusion process, which is based on thermodynamics. The process involves gradually adding noise to a sample from the data distribution, and then training a neural network to reverse this process by gradually removing the noise. Recent developments in diffusion models have focused on generative tasks, \eg, image generation~\cite{ho2020denoising,song2020denoising,dhariwal2021diffusion,ho2022classifier,wang2022zero}, natural language generation~\cite{austin2021structured,li2022diffusion,gong2022diffuseq}, and audio generation~\cite{popov2021grad}. Some other works have attempted to adapt the diffusion model for discriminant tasks, \eg, image segmentation~\cite{amit2021segdiff,baranchuk2021label,brempong2022denoising}, visual grounding~\cite{cheng2023parallel}, and detection~\cite{chen2022diffusiondet}. However, there are no previous works that adapt generative diffusion models for cross-modal retrieval. This work addresses this gap by modeling the correlation between text and video as their joint probability and utilizing the diffusion model to gradually generate the joint probability distribution from noise. To the best of our knowledge, we are the first to adapt the diffusion model for cross-modal retrieval.

\section{Method}
In this paper, we tackle the tasks of text-to-video and video-to-text retrieval. In the task of text-to-video retrieval, we are given a text query $t$ and a gallery of videos $\bm{V}$. The goal is to rank all videos $v\in \bm{V}$ so that the video corresponding to the text query $t$ is ranked as high as possible. Similarly, the goal of video-to-text retrieval is to rank all text candidates $t\in \bm{T}$ based on the video query $v$.

\subsection{Existing Solutions: Discriminant Modeling}
A common method for the retrieval problem is similarity learning~\cite{xing2002distance}. Specifically, the caption as well as the video are represented in a common multi-dimensional embedding space, where the similarity can be calculated as the dot product of their corresponding representations. In the task of text-to-video retrieval, such methods~\cite{luo2021clip4clip,gorti2022x,jin2022expectation} use the logits to calculate the posterior probability:
\begin{equation}
p(v|t;\bm{\theta^{t}},\bm{\theta^{v}})\!=\!\frac{\exp \big(f_{\bm{\theta^{t}}}(t)^\top f_{\bm{\theta^{v}}}(v)/\tau \big)}{\sum_{v^{'}\in \bm{V}}\exp \big(f_{\bm{\theta^{t}}}(t)^\top f_{\bm{\theta^{v}}}(v^{'})/\tau \big)},
\end{equation}
where $\tau$ is the temperature hyper-parameter. $\bm{\theta_{t}}$ and $\bm{\theta_{v}}$ are parameters of the text feature extractor and video feature extractor, respectively. Finally, existing methods rank all video candidates based on the posterior probability $p(v|t;\bm{\theta_{t}},\bm{\theta_{v}})$. Similarly, in the video-to-text retrieval, they rank all text candidates based on $p(t|v;\bm{\theta_{t}},\bm{\theta_{v}})$.
The parameters $\{\bm{\theta_{t}}, \bm{\theta_{v}}\}$ of text feature extractor and video feature extractor are optimized by minimizing the contrastive learning loss~\cite{oord2018representation,zhang2021zero,zhang2022align,zhang2023patch}:
\begin{equation}
\begin{aligned}
\bm{\theta_{t}^{*}}, \bm{\theta_{v}^{*}}=\mathop{\arg\min}\limits_{\bm{\theta_{t}}, \bm{\theta_{v}}} -\frac{1}{2} \mathbb{E}_{(t,v)\in \mathcal{D}} \big [ \log p(v|t;\bm{\theta_{t}},\bm{\theta_{v}})\\ + \log p(t|v;\bm{\theta_{t}},\bm{\theta_{v}}) \big ],
\end{aligned}
\end{equation}
where $\mathcal{D}$ is a corpus of text-video pairs $(t,v)$. Such learning strategy is equivalent to maximizing conditional likelihood, \ie, $\prod_{(t,v)\in \mathcal{D}}p(v|t)+\prod_{(t,v)\in \mathcal{D}}p(t|v)$, which is called \textit{discriminative training}~\cite{bernardo2007generative}. 

Since existing methods directly model the conditional probability distribution $p(v|t)+p(t|v)$, without considering the input distribution $p(t)$ and $p(v)$, they fail to achieve good generalization on unseen data.

\subsection{DiffusionRet: Generation Modeling}
Our method reformulates the retrieval task from a generative modeling perspective. As Feynman’s mantra ``what I cannot create, I do not understand'' shows, we argue that it is time to rethink the current discriminative retrieval regime.

Inspired by the great success of diffusion models~\cite{sohl2015deep,ho2020denoising,dhariwal2021diffusion}, we adopt the diffusion model as the generator. Concretely, given a query and $N$ candidates, our goal is to synthesize the distribution $x^{1:N}=\{x^i\}_{i=1}^{N}$ from Gaussian noise $\mathcal{N}(0,\bm{\textrm{I}})$. In contrast to the prior works, which typically optimize the posterior probabilities $p(v|t;\bm{\theta_{t}},\bm{\theta_{v}})+p(t|v;\bm{\theta_{t}},\bm{\theta_{v}})$, our method builds the joint probabilities:
\begin{equation}
\begin{aligned}
x^{1:N}=p(v, t|\bm{\phi})=f_{\bm{\phi}}\big(v, t, \mathcal{N}(0,\bm{\textrm{I}})\big),
\end{aligned}
\end{equation}
where $\bm{\phi}$ is the parameter of the generator. It is worth noting that the learning objective of the generator is equivalent to approximating the data distribution, \ie, $\prod_{(t,v)\in \mathcal{D}}p(v, t)$, which is called \textit{generative training}~\cite{bernardo2007generative}. The overview of our DiffusionRet is shown in Fig.~\ref{fig2}.

\subsubsection{{Text-Frame Attention Encoder}}
To extract the joint encoding of text and video, we propose the text-frame attention encoder, which takes text representation as query and frame representation as key and value.

Specifically, for text representation, we adopt the text encoder of CLIP (ViT-B/32)~\cite{radford2021learning} and take the output from the [CLS] token as the text representation $C_t\in \mathbb{R}^D$, where $D$ is the size of the dimension. For video frame representation, we extract the frames evenly from the video clip as the input frame sequence. Then we use ViT~\cite{dosovitskiy2021an} to encode the frame sequence and adapt the output from the [CLS] token as the frame embedding. After extracting the frame embedding, we use a 4-layer transformer to aggregate the embedding of all frames and obtain the frame representation $F$. 

Then, we aggregate frame representation with the text-frame attention encoder. Concretely, we feed the text representation $C_t\in \mathbb{R}^D$ as query and the frame representation $F$ as key and value into an attention module. The final video representation is defined as:
\begin{equation}
C_v=\textrm{Softmax}(C_tF^\top/{{\tau}^{'}})F,
\end{equation}
where ${\tau}^{'}$ is the trade-off hyper-parameter. The smaller ${\tau}^{'}$ allows visual features to take more textual information into account when aggregated.

\subsubsection{Query-Candidate Attention Denoising Network}
\begin{figure}[tbp]
\centering
\includegraphics[width=1.\linewidth]{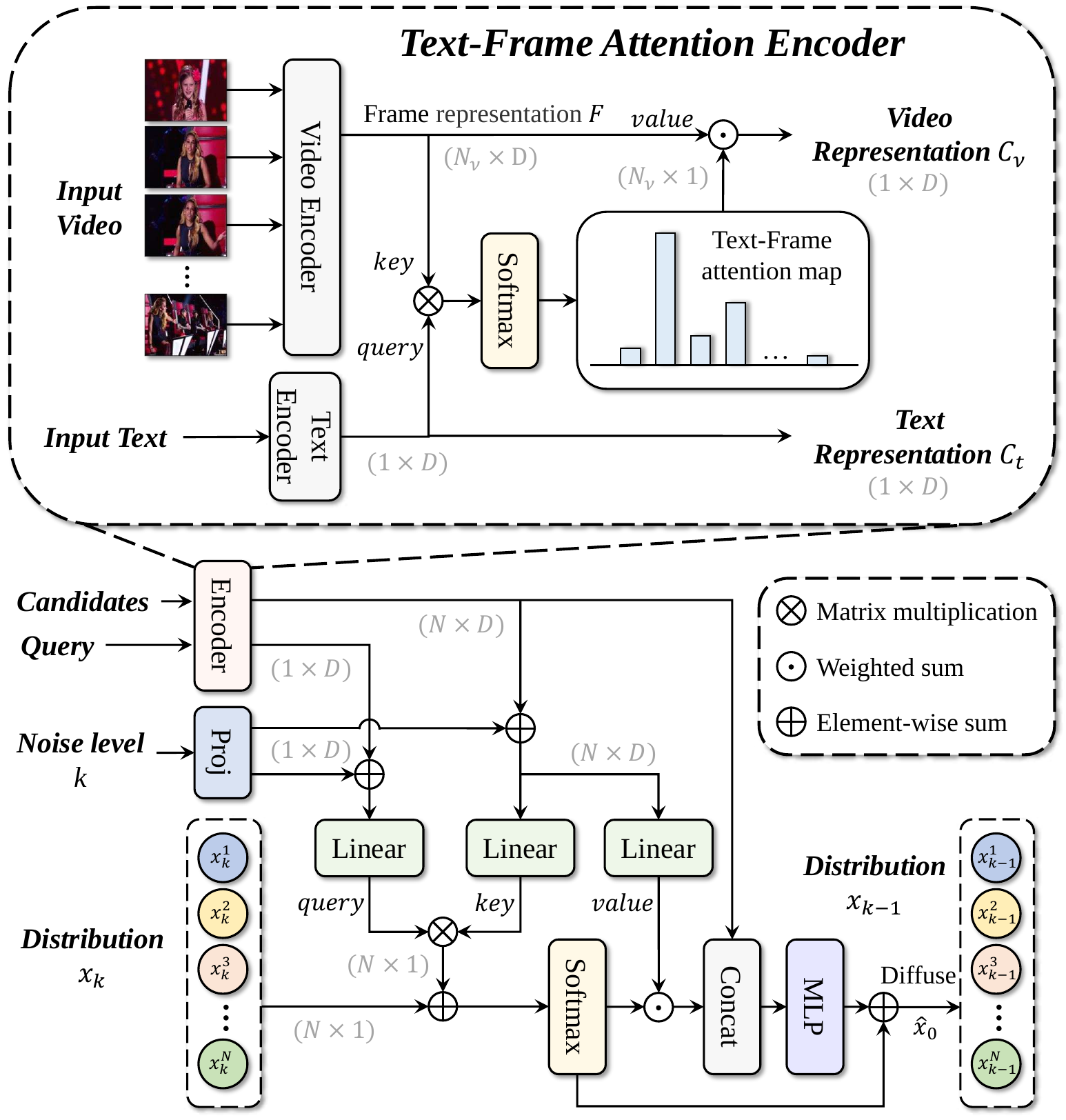}
\vspace{-1.5em}
\caption{\textbf{The model architecture of the denoising network.} We first leverage the text-frame attention encoder to extract the joint encoding of text and video. Then, we feed a distribution $x_k$ of length $N$, as well as noise level $k$, and the text and video representations into the query-candidate attention network. In each sampling step, the denoising network predicts the final clean distribution $\hat{x}_0$.}
\label{fig3}
\end{figure}

Different from other generation tasks which only focus on the authenticity and diversity of generation, the key to the retrieval task is to mine the correspondence between the query and candidates. To this end, we propose the query-candidate attention denoising network to capture the correspondence between query and candidates in the generation process. The overview of the denoising network is shown in Fig.~\ref{fig3}.

We start with the task of text-to-video retrieval. To elaborate, we first project the text representation $C_t\in \mathbb{R}^D$ into the query and video representation $C_v\in \mathbb{R}^{N\times D}$ into key and value, where $N$ is the number of video candidates. The projections are formulated as:
\begin{equation}
\begin{aligned}
&Q_t=W_{Q_t}\big(C_t+\textrm{Proj}(k)\big),\\
&K_v=W_{K_v}\big(C_v+\textrm{Proj}(k)\big),\\
&V_v=W_{V_v}\big(C_v+\textrm{Proj}(k)\big),
\end{aligned}
\end{equation}
where $W_{Q_t}$, $W_{K_v}$ and $W_{V_v}$ are projection matrices. ``Proj'' projects the noise level $k$ into $D$ dimensional embedding. To give more weight to the video candidates with higher joint probabilities of the previous noise level, we add the distribution $x_{k}$ to the attention weight. The attention mechanism can be defined as:
\begin{equation}
E_t=\textrm{Softmax}({Q_tK_v^\top}+x_{k})V_v.
\end{equation}
We treat the output $E_t$ of the attention module as high semantic level embedding which contains textual query information. Then, we concatenate video representation $C_v$ and embedding $E_t$, generating the input data of the denoising decoder $[C_v, E_t]\in \mathbb{R}^{N\times 2D}$. The denoising decoder is a multi-layer perceptron (MLP) containing a linear layer with a Relu activation function~\cite{glorot2011deep} for encoding features and a linear layer for calculating output distribution.

Similarly, in the video-to-text retrieval, we feed the projection of video representation as query and the projection of text representation as key and value into an attention module. The output distribution is calculated in the same way.

\begin{figure*}[tbp]
\centering
\includegraphics[width=1.\linewidth]{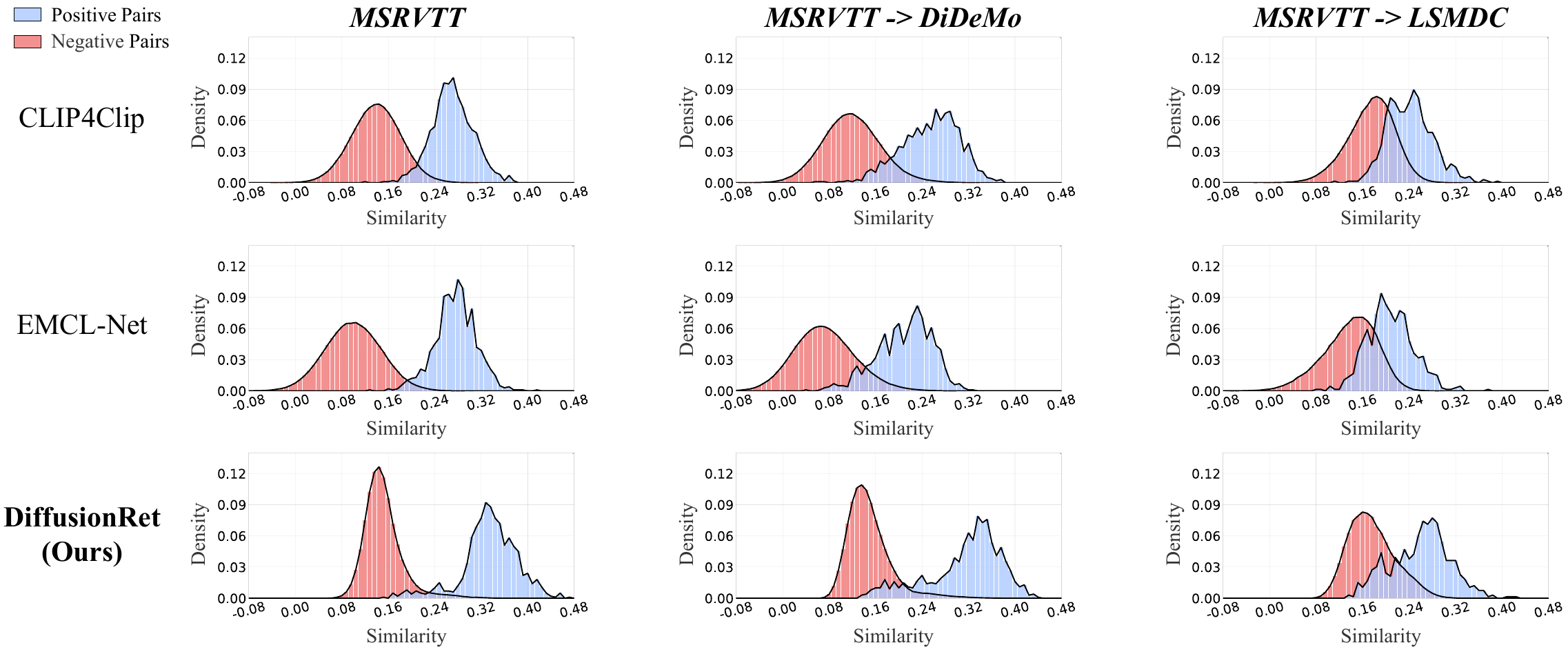}
\caption{\textbf{Similarity distribution for positive and negative pairs in in-domain retrieval and out-domain retrieval.} ``\textit{MSRVTT}'' represents the similarity distribution on the MSRVTT test setting using pre-trained models on the MSRVTT dataset. ``\textit{A}-\textgreater{}\textit{B}'' indicates that ``\textit{A}'' is the source domain and ``\textit{B}'' is the target domain. For example, ``\textit{MSRVTT}-\textgreater{}\textit{DiDeMo}'' denotes that the generalization results on unseen DiDeMo test setting using pre-trained models on the MSRVTT dataset.}
\label{fig:p}
\end{figure*}

\subsubsection{Optimization from both Generation Perspective and Discrimination Perspective}\label{Optimization}
Compared to the generation methods, discriminant methods usually give good predictive performance~\cite{bernardo2007generative}. To leverage the benefit of both generative and discriminative methods, we optimize the proposed generation model from both generation and discrimination perspectives.

\textbf{Probabilistic Diffusion (Generation Perspective).} In the generation perspective, we model the distribution $x = p(v, t|\bm{\phi})$ as the reversed diffusion process of the $K$-length Markov Chain. Specifically, our method learns the joint distribution of text and video by gradually denoising a variable sampled from Gaussian distribution. In a forward diffusion process $q(x_k|x_{k-1})$, noised sampled from Gaussian distribution is added to a ground truth data distribution $x_0$ at every noise level $k$:
\begin{equation}
\begin{aligned}
q(x_k|x_{k-1})&=\mathcal{N}(x_k;\sqrt{1-\beta_{k}}x_{k-1},\beta_{k}\bm{\textrm{I}}),\\
q(x_{1:K}|x_{0})&=\prod_{k=1}^{K}q(x_k|x_{k-1}),
\end{aligned}
\label{Markov}
\end{equation}
where ${\beta}_k$ decides the noise schedule which gradually increases. We can sample $x_k$ by the following formula:
\begin{equation}
x_k = \sqrt{\bar{\alpha}_{k}}x_0 + \sqrt{1-\bar{\alpha}_{k}}\epsilon,
\end{equation}
where $\bar{\alpha}_{k}=\prod_{i=1}^{k}\alpha_i$ and ${\alpha}_{k}=1-{\beta}_{k}$. $\epsilon$ is a noise sampled from $\mathcal{N}(0,1)$. Instead of predicting the noise component $\epsilon$, we follow \cite{ramesh2022hierarchical} and predict the data distribution itself, \ie, $\hat{x}_0=f_{\bm{\phi}}(v, t, x_k)$. The training objective of the diffusion model can be defined as:
\begin{equation}
\mathcal{L}_{G} = \mathbb{E}_{(t,v)\in \mathcal{D}} \Big [\textrm{KL}\big(x_0\|f_{\bm{\phi}}(v, t, x_k)\big) \Big ].
\end{equation}
This loss maximizes the likelihood of $p(v,t)$ by bringing $f_{\bm{\phi}}(v, t, x_k)$ and $x_0$ closer together.

\newcommand{\pub}[1]{\color{gray}{\tiny{#1}}}
\newcommand{\Frst}[1]{{\textbf{#1}}}
\newcommand{\Scnd}[1]{{{#1}}}
\newcommand{\Mul}[1]{\color{gray}{#1}}

\begin{table*}[t]
\footnotesize
\centering
\resizebox{1.\linewidth}{!}{
\begin{tabular}{l|cccccc|cccccc}
\toprule[1.25pt]
\multirow{2}{*}{\textbf{Methods}} &\multicolumn{6}{c|}{\textbf{Text-to-Video}} &\multicolumn{6}{c}{\textbf{Video-to-Text}}\\
\cmidrule(rl){2-7}\cmidrule(rl){8-13}
  & R@1$\uparrow$ & R@5$\uparrow$ & R@10$\uparrow$ & Rsum$\uparrow$ & MdR$\downarrow$ & MnR$\downarrow$ & R@1$\uparrow$ & R@5$\uparrow$ & R@10$\uparrow$ & Rsum$\uparrow$ & MdR$\downarrow$ & MnR$\downarrow$ \\ \midrule
MMT~{\cite{gabeur2020multi}}~\pub{ECCV20} & {26.6} & 57.1 & 69.6 & 153.3 & 4.0 & 24.0 & 27.0 & 57.5 & 69.7 & 154.2 & 3.7 & 21.3 \\
Support-Set~{\cite{patrick2021support}}~\pub{ICLR21}  & {30.1} & 58.5 & 69.3 & 157.9 & \Scnd{3.0} & - &28.5 &58.6 &71.6 & 158.7 &\Scnd{3.0} & - \\
T2VLAD~{\cite{wang2021t2vlad}}~\pub{CVPR21} & {29.5} & 59.0 & 70.1 & 158.6 & 4.0 & -  & 31.8 &60.0 &71.1 & 162.9 &\Scnd{3.0} & -\\
TT-CE~{\cite{croitoru2021teachtext}}~\pub{ICCV21} & {29.6} & 61.6 & 74.2 & 165.4 & \Scnd{3.0} & - & 32.1 & 62.7 & 75.0 & 169.8 &\Scnd{3.0} & -\\
FROZEN~{\cite{bain2021frozen}}~\pub{ICCV21} & {31.0} & 59.5 & 70.5 & 161.0 & \Scnd{3.0} & - & - &- &- &- & - & -\\
CLIP4Clip~{\cite{luo2021clip4clip}}~\pub{Neurocomputing22} & {44.5} & 71.4 & 81.6 & 197.5 &\Frst{2.0} & 15.3 & 42.7 & 70.9& 80.6 & 194.2 & \Frst{2.0} & 11.6\\
X-Pool~{\cite{gorti2022x}}~\pub{CVPR22}  & {46.9} & 72.8 & 82.2 & 201.9 &\Frst{2.0} & 14.3 & 44.4 & 73.3 & 84.0 & 201.7 & \Frst{2.0} & 9.0 \\
TS2-Net~{\cite{liu2022ts2}}~\pub{ECCV22}  & {47.0} & \Scnd{74.5} & \Frst{83.8} &{205.3} & \Frst{2.0} & \Scnd{13.0} & 45.3 & 74.1 & 83.7 & 203.1 & \Frst{2.0} & 9.2 \\
EMCL-Net~{\cite{jin2022expectation}}~\pub{NeurIPS22}  & {46.8} & 73.1 &\Scnd{83.1} & 203.0 &\Frst{2.0} & 12.8 & 
46.5 & 73.5 & 83.5 & 203.5 & \Frst{2.0} & 8.8 \\
\Mul{TABLE}~\cite{chen2023tagging}\ssymbol{2}~\pub{AAAI23} &\Mul{47.1} &\Mul{74.3} &\Mul{82.9} &\Mul{204.3} &\Mul{{2.0}} &\Mul{13.4} &\Mul{47.2} &\Mul{\Scnd{74.2}} &\Mul{{84.2}} &\Mul{205.6} &\Mul{{2.0}} &\Mul{11.0} \\
DiCoSA~{\cite{jin2023text}}~\pub{IJCAI23} & 47.5 & 74.7 & \textbf{83.8} & 206.0 & \textbf{2.0} & 13.2 & 46.7 & \textbf{75.2} & 84.3 & 206.2 & \textbf{2.0} & 8.9 \\
HBI~\cite{jin2022video}~\pub{CVPR23} & 48.6 & 74.6 & 83.4 & 206.6 & \textbf{2.0} & \textbf{12.0} & 46.8 & {74.3} & 84.3 & 205.4 & \textbf{2.0} & 8.9 \\
\midrule
\rowcolor{aliceblue!60} \textbf{DiffusionRet (Ours)} & \textbf{49.0} & \textbf{75.2} & 82.7 & \Scnd{206.9} & \textbf{2.0} &{12.1} & \Scnd{47.7} & 73.8 &\Frst{84.5} & \Scnd{206.0} &\textbf{2.0} &\Scnd{8.8} \\
\rowcolor{aliceblue!60} + QB-Norm~\cite{bogolin2022cross} & \Scnd{48.9} & \textbf{75.2} & \Scnd{83.1} & \Frst{207.2} & \textbf{2.0} &{12.1} & \textbf{49.3} &{74.3} & 83.8 & \Frst{207.4} &\textbf{2.0} & \Frst{8.5} \\
\bottomrule[1.25pt]
\end{tabular}
}
\caption{\textbf{Comparisons to current state-of-the-art methods on the MSRVTT dataset.} ``$\uparrow$'' denotes that higher is better. ``$\downarrow$'' denotes that lower is better. ``\ssymbol{2}'' denotes that the model uses additional expert features, \eg, TABLE uses the object, person, scene, motion, and audio features from multiple pretrained experts. We gray out TABLE for a fair comparison.}
\label{MSRVTT results}
\end{table*}

\begin{table*}[htb]
\centering
\footnotesize
\resizebox{1.\linewidth}{!}
{
\begin{tabular}{l|cccccc|cccccc}
\toprule[1.25pt]
\multirow{2}{*}{\textbf{Method}} &  \multicolumn{6}{c|}{\textbf{LSMDC}}        & \multicolumn{6}{c}{\textbf{MSVD}}  \\ 
\cmidrule(rl){2-7}\cmidrule(rl){8-13}
  & R@1$\uparrow$ & R@5$\uparrow$ & R@10$\uparrow$ & Rsum$\uparrow$ & MdR$\downarrow$ & MnR$\downarrow$
& R@1$\uparrow$ & R@5$\uparrow$ & R@10$\uparrow$ & Rsum$\uparrow$ & MdR$\downarrow$ & MnR$\downarrow$    \\ \midrule
FROZEN~{\cite{bain2021frozen}}~\pub{ICCV21} & 15.0 & 30.8 & 39.8 & 85.6 & 20.0 & - & 33.7 & 64.7 & 76.3 & 174.7 & 3.0 & -\\
TT-CE~{\cite{croitoru2021teachtext}}~\pub{ICCV21} & {17.2} & 36.5 & 46.3 & 100.0 & 13.7 & - & 25.4 & 56.9 & 71.3 & 153.6 & 4.0 & -\\
CLIP4Clip~{\cite{luo2021clip4clip}}~\pub{Neurocomputing22}  & 22.6 & 41.0 & 49.1 & 112.7 & 11.0 & 61.0 & 45.2 & 75.5 & {84.3} & 205.0 & \Frst{2.0} & \textbf{10.3} \\
TS2-Net~{\cite{liu2022ts2}}~\pub{ECCV22}  & 23.4 & 42.3 & 50.9 & 116.6 & 9.0 & 56.9 & - & - & - & - & - & -\\
EMCL-Net~{\cite{jin2022expectation}}~\pub{NeurIPS22}  & 23.9 & 42.4 & 50.9 & 117.2 & 10.0 & - & 42.1 & 71.3 & 81.1 & 194.5 & \textbf{2.0} & 17.6 \\
\midrule
\rowcolor{aliceblue!60} \textbf{DiffusionRet (Ours)}  & \Frst{24.4} & \Frst{43.1} & \Frst{54.3} & \textbf{121.8} & \Frst{8.0} & \Frst{40.7} & {46.6} & {75.9} & 84.1 & 206.6 & \Frst{2.0} & 15.7      \\
\rowcolor{aliceblue!60} + QB-Norm~\cite{bogolin2022cross} & {23.9} & {42.7} & {53.6} & {120.2} & \textbf{8.0} & \textbf{40.7} & \textbf{47.9} &\textbf{77.2} &\textbf{84.8} &\textbf{209.9} & \textbf{2.0} & 15.6 \\
\midrule[1.pt]
\multirow{2}{*}{\textbf{Method}} &  \multicolumn{6}{c|}{\textbf{ActivityNet Captions}}        & \multicolumn{6}{c}{\textbf{DiDeMo}}  \\ 
\cmidrule(rl){2-7}\cmidrule(rl){8-13}
  & R@1$\uparrow$ & R@5$\uparrow$ & R@10$\uparrow$ & Rsum$\uparrow$ & MdR$\downarrow$ & MnR$\downarrow$
& R@1$\uparrow$ & R@5$\uparrow$ & R@10$\uparrow$ & Rsum$\uparrow$ & MdR$\downarrow$ & MnR$\downarrow$    \\ \midrule
ClipBERT~{\cite{lei2021less}}~\pub{CVPR21} & 21.3 & 49.0 & 63.5 & 133.8 & 6.0 & - & 20.4 & 48.0 & 60.8 & 129.2 & 6.0 & - \\
CLIP4Clip~{\cite{luo2021clip4clip}}~\pub{Neurocomputing22}  & 40.5 & 72.4 & 83.6 & 196.5 & \Frst{2.0} & 7.5 & 42.8 & 68.5 & 79.2 & 190.5 & \Frst{2.0} & 18.9 \\
TS2-Net~{\cite{liu2022ts2}}~\pub{ECCV22}  & 41.0 & {73.6} & 84.5 & 199.1 & \Frst{2.0} & 8.4 & 41.8 & 71.6 & 82.0 & 195.4 & \Frst{2.0} & 14.8 \\
EMCL-Net~{\cite{jin2022expectation}}~\pub{NeurIPS22}  & 41.2 & 72.7 & 83.6 & 197.5 & \textbf{2.0} & 8.6 & 45.3 & 74.2 & 82.3 & 201.8 & \textbf{2.0} & 12.3 \\
DiCoSA~{\cite{jin2023text}}~\pub{IJCAI23} & 42.1 & 73.6 & 84.6 & 200.3 & \textbf{2.0} & 6.8 & 45.7 & 74.6 & \textbf{83.5} & 203.8 & \textbf{2.0} & \textbf{11.7} \\
HBI~\cite{jin2022video}~\pub{CVPR23} & 42.2 & 73.0 & 84.6 & 199.8 & \textbf{2.0} & 6.6 & 46.9 & 74.9 & 82.7 & 204.5 & \textbf{2.0} & 12.1 \\
\midrule
\rowcolor{aliceblue!60} \textbf{DiffusionRet (Ours)}  & {45.8} & \textbf{75.6} & \textbf{86.3} & {207.7} &\textbf{2.0} & \textbf{6.5} & {46.7} & {74.7} & {82.7} & 204.1 &\textbf{2.0} & {14.3}        \\ 
\rowcolor{aliceblue!60} + QB-Norm~\cite{bogolin2022cross} & \textbf{48.1} & \textbf{75.6} & {85.7} & \textbf{209.4} &\textbf{2.0} & {6.8} & \textbf{48.9} &\textbf{75.5} &{83.3} & \textbf{207.7} &\textbf{2.0} & {14.1} \\ 
\bottomrule[1.25pt]
\end{tabular}
}
\caption{\textbf{Text-to-video retrieval performance on other datasets.} ``$\uparrow$'' denotes that higher is better. ``$\downarrow$'' denotes that lower is better. Video-to-text retrieval performance is provided in the supplementary material.}
\label{Experiments0}
\end{table*}

\myparagraph{Contrastive Learning (Discrimination Perspective).} In the discrimination perspective, we optimize the features which are input into the generator so that these features contain discriminant semantic information. Inspired by~\cite{wang2022disentangled}, we align representations of text and video at the token level. Specifically, we take all tokens output by the text encoder as word-level features $\{w^{i}\}^{N_t}_{i=1}$, where $N_t$ is the length of the text. Frame-level features $\{f^{j}\}^{N_v}_{j=1}$ are all tokens output by the video encoder, where $N_v$ is the length of the video. Then, we calculate the alignment matrix $A=[a_{ij}]^{N_{t}\times N_{v}}$, where $a_{ij}=\frac{(w^{i})^\top f^{j}}{\Vert w^{i}\Vert \Vert f^{j}\Vert}$ is the cosine similarity between the $i_{th}$ word and the $j_{th}$ frame. The total similarity score consists of two parts: text-to-video similarity and video-to-text similarity. For the text-to-video similarity, we first calculate the maximum alignment score of the $i_{th}$ word as $\underset{j}{\textrm{max}}\ a_{ij}$. We then take the weighted average maximum alignment score over all words. For the video-to-text similarity, we take the weighted average maximum alignment score over all frames. The total similarity score can be defined as: 
\begin{equation}
s=\frac{1}{2}(\underbrace{\sum_{i=1}^{N_t} g_t^i\ \underset{j}{\textrm{max}}\ a_{ij}}_{\textrm{text-to-video similarity}}+\underbrace{\sum_{j=1}^{N_v} g_v^j\ \underset{i}{\textrm{max}}\ a_{ij}}_{\textrm{video-to-text similarity}}),
\end{equation}
where $\{g_t^{i}\}_{i=1}^{N_t}\!=\!\textrm{Softmax}\big(\textrm{MLP}_t(\{w^{i}\}_{i=1}^{N_t})\big)$ and $\{g_v^{j}\}_{j=1}^{N_v}\!\!=\!\!\textrm{Softmax}\big(\textrm{MLP}_v(\{f^{j}\}_{j=1}^{N_v})\big)$ are the weights of the text words and video frames, respectively. Then, the contrastive loss~\cite{oord2018representation} can be formulated as:
\begin{equation}
\begin{aligned}
\mathcal{L}_{D}=-\frac{1}{2}\mathbb{E}_{(t,v)\in \mathcal{D}}\big [ \log \frac{\exp(s_{t,v}/\hat{\tau})}{\sum_{v^{'}\in \bm{V}}\exp(s_{t,v^{'}}/\hat{\tau})}\\
+\log \frac{\exp(s_{t,v}/\hat{\tau})}{\sum_{t^{'}\in \bm{T}}\exp(s_{t^{'},v}/\hat{\tau})} \big ],
\end{aligned}
\end{equation}
where $s_{t,v}$ is the similarity score between text $t$ and video $v$. $\hat{\tau}$ is the temperature hyper-parameter. This loss brings semantically similar texts and videos closer together in the representation space, thus helping the diffusion model to generate the joint distribution of text and video.

\begin{table*}[t]
\centering
\footnotesize
\setlength{\tabcolsep}{3.pt}
{
\subfloat[\textbf{Effect of the generation loss type.}]
{
\begin{tabular}{lcccc}
\toprule[1.25pt]
Loss type & R@1$\uparrow$ & R@5$\uparrow$ & R@10$\uparrow$ & MnR$\downarrow$ \\ \midrule
MSE & 46.9 & 75.0 & 82.5 & 12.2 \\
\rowcolor{aliceblue!60} KL & \Frst{49.0} & \Frst{75.2} & \Frst{82.7} & \Frst{12.1} \\
\bottomrule[1.25pt]
\end{tabular}
\label{tab:loss}
}
\quad \
\subfloat[\textbf{Effect of the sampling strategy.}]
{
\begin{tabular}{lcccc}
\toprule[1.25pt]
Sampling  & R@1$\uparrow$ & R@5$\uparrow$ & R@10$\uparrow$ & MnR$\downarrow$ \\ \midrule
DDPM & 48.9 & \Frst{75.2} & 82.6 & \Frst{12.1} \\
\rowcolor{aliceblue!60} DDIM & \Frst{49.0} & \Frst{75.2} & \Frst{82.7} & \Frst{12.1} \\
\bottomrule[1.25pt]
\end{tabular}
\label{tab:sampling}
}
\quad \
\subfloat[\textbf{Effect of the schedule of $\beta$.}]
{
\begin{tabular}{lcccc}
\toprule[1.25pt]
Schedule of $\beta$  & R@1$\uparrow$ & R@5$\uparrow$ & R@10$\uparrow$ & MnR$\downarrow$ \\ \midrule
Linear & 48.7 & 75.1 & \Frst{82.8} & 12.3 \\
\rowcolor{aliceblue!60} Cosine & \Frst{49.0} & \Frst{75.2} & 82.7 & \Frst{12.1} \\
\bottomrule[1.25pt]
\end{tabular}
\label{tab:schedule}
} 
\\ 
\subfloat[\textbf{Effect of the training strategy.}]
{
\begin{tabular}{lcccc}
\toprule[1.25pt]
Training strategy  & R@1$\uparrow$ & R@5$\uparrow$ & R@10$\uparrow$ & MnR$\downarrow$ \\ \midrule
CLIP zero-shot & 25.3 & 46.5 & 56.0 & 75.8\\ \midrule
\underline{Gen}eration (Gen) & 40.2 & 68.6 & 78.8 & 16.1 \\
\underline{Dis}crimination (Dis) & 46.8 & 73.5 & \Frst{82.9} & 13.0 \\
\rowcolor{aliceblue!60} Both Gen and Dis  & \Frst{49.0} & \Frst{75.2} & 82.7 & \Frst{12.1}  \\
\bottomrule[1.25pt]
\end{tabular}
\label{tab:training}
}
\quad \
\subfloat[\textbf{{Effect of the diffusion steps on R@1.}}]
{
\begin{tabular}{lcccc}
\toprule[1.25pt]
\diagbox{{train}}{{eval}}  & 10 & 50 & 100 & 1000 \\ \midrule
10 & 48.7 & {\color{gray}\ding{56}} & {\color{gray}\ding{56}} & {\color{gray}\ding{56}} \\
50 & 48.6 & \colorbox{aliceblue!60}{\Frst{49.0}} & {\color{gray}\ding{56}} & {\color{gray}\ding{56}} \\
100 & 48.1 & 48.5 & 48.6 & {\color{gray}\ding{56}} \\
1000 & 48.2 & 48.4 & 48.5 & 48.6 \\
\bottomrule[1.25pt]
\end{tabular}
\label{tab:step}
}
\quad \
\subfloat[\textbf{Effect of the scale of $\beta$.}]
{
\begin{tabular}{lcccc}
\toprule[1.25pt]
Scale of $\beta$  & R@1$\uparrow$ & R@5$\uparrow$ & R@10$\uparrow$ & MnR$\downarrow$ \\ \midrule
0.1 & 48.4 & \Frst{75.3} & \Frst{82.7} & 12.1 \\
0.5 & 48.5 & 75.1 & 82.6 & \Frst{12.0} \\
\rowcolor{aliceblue!60} 1.0 & \Frst{49.0} & 75.2 & \Frst{82.7} & 12.1 \\
1.5 & 48.3 & \Frst{75.3} & \Frst{82.7} & 12.2 \\
2.0 & 48.5 & 75.2 & 82.5 & 12.1 \\
\bottomrule[1.25pt]
\end{tabular}
\label{tab:scale}
}
\vspace{-0.5em}
\caption{\textbf{Ablation study on the MSRVTT dataset.} {``KL'' denotes Kullback-Leibler divergence. ``Gen'' denotes generation. ``Dis'' denotes discrimination. ``{\color{gray}\ding{56}}'' denotes that the evaluation process fails. Default settings are marked in \colorbox{aliceblue!60}{blue}.}}
\label{Comparisons to State-of-the-arts}
}
\end{table*}

\section{Experiments}
\subsection{Experimental Settings}
\myparagraph{Datasets.} 
\textbf{MSRVTT}~\cite{xu2016msr} contains 10,000 YouTube videos, each with 20 text descriptions. We follow the 1k-A split~\cite{liu2019use} with 9,000 videos for training and 1,000 for testing. \textbf{LSMDC}~\cite{rohrbach2015a} contains 118,081 video clips from 202 movies. We follow the split of \cite{gabeur2020multi} with 1,000 videos for testing. \textbf{MSVD}~\cite{chen2011collecting} contains 1,970 videos. We follow the official split of 1,200 and 670 as the train and test set, respectively. \textbf{ActivityNet Captions}~\cite{krishna2017dense} contains 20,000 YouTube videos. We report results on the ``val1'' split of 10,009 and 4,917 as the train and test set. \textbf{DiDeMo}~\cite{anne2017localizing} contains 10,464 videos annotated 40,543 text descriptions. We follow the training and evaluation protocol in \cite{luo2021clip4clip}.

\myparagraph{Metrics.} We choose Recall at rank K (R@K), the Sum of Recall at rank $\{1,5,10\}$ (Rsum), Median Rank (MdR), and mean rank (MnR) to evaluate the retrieval performance.

\begin{table*}[t]
\footnotesize
\centering
\resizebox{1.\linewidth}{!}
{
\begin{tabular}{lc|ccc|ccc|ccc}
\toprule[1.25pt]
\multirow{3}{*}{\textbf{Methods}} & \multirow{3}{*}{\textbf{Steps}} & \multicolumn{3}{c|}{\textbf{Candidate gallery size is 1,000}} & \multicolumn{3}{c|}{\textbf{Candidate gallery size is is 3,000}} & \multicolumn{3}{c}{\textbf{Candidate gallery size is 5,000}}\\
\cmidrule(rl){3-5}\cmidrule(rl){6-8}\cmidrule(rl){9-11}
 &  &{Inference} &{GPU} &\multirow{2}{*}{{R@1$\uparrow$}} &{Inference} &{GPU} &\multirow{2}{*}{{R@1$\uparrow$}} &{Inference} &{GPU} &\multirow{2}{*}{{R@1$\uparrow$}} \\
 & & Time (s)$\downarrow$ & Memory (M)$\downarrow$ & & Time (s)$\downarrow$ & Memory (M)$\downarrow$ &  & Time (s)$\downarrow$ & Memory (M)$\downarrow$ & \\
 \midrule
CLIP4Clip~{\cite{luo2021clip4clip}} & {\color{gray}\ding{56}} & 53.20 & 3043 & 43.3 & 137.24 & 3095 & 37.5 & 244.63 & 3147 & 32.6 \\
X-Pool~{\cite{gorti2022x}} & {\color{gray}\ding{56}} & 63.02 & 5585 & 45.9 & 330.02 & 5587 & 31.3 & 833.11 & 5590 & 27.0 \\
TS2-Net~{\cite{liu2022ts2}} & {\color{gray}\ding{56}} & 117.71 & 2803 & 46.5 & 702.65 & 3053 & 37.8 & 1866.18 & 3303 & 32.8 \\
\midrule
\rowcolor{aliceblue!60} \textbf{DiffusionRet (Ours)} & 1 & 59.02 & 3253 & 23.4 & 143.04 & 3747 & 16.6 & 260.67 & 4297 & 13.1 \\
\rowcolor{aliceblue!60} \textbf{DiffusionRet (Ours)} & 3 & 59.32 & 3253 & 47.8 & 144.42 & 3747 & 37.8 & 260.96 & 4297 & 30.5 \\
\rowcolor{aliceblue!60} \textbf{DiffusionRet (Ours)} & 5 & 59.55 & 3253 & 47.9 & 144.63 & 3747 & 37.8 & 261.11 & 4297 & 32.9 \\
\rowcolor{aliceblue!60} \textbf{DiffusionRet (Ours)} & 10 & 59.70 & 3253 & 48.6 & 145.97 & 3747 & 37.9 & 261.69 & 4297 & 33.0 \\
\rowcolor{aliceblue!60} \textbf{DiffusionRet (Ours)} & 30 & 60.66 & 3253 & 48.9 & 148.11 & 3747 & 38.1 & 274.49 & 4297 & 33.1 \\
\rowcolor{aliceblue!60} \textbf{DiffusionRet (Ours)} & 50 & 62.04 & 3253 & \textbf{49.0} & 151.31 & 3747 & \textbf{38.2} & 281.27 & 4297 & \textbf{33.2} \\ 
\bottomrule[1.25pt]
\end{tabular}
}
\caption{\textbf{Evaluation of the Inference costs with a single RTX3090 GPU.} We report the R@1 metric for the text-to-video task and repeat the experiment three times to take the average value. ``Steps'' denotes the diffusion steps. ``{\color{gray}\ding{56}}'' denotes that this method does not apply this parameter. ``$\uparrow$'' denotes that higher is better. ``$\downarrow$'' denotes that lower is better.}
\label{tab:cost}
\end{table*}

\begin{table*}[t]
\footnotesize
\centering
\resizebox{1.\linewidth}{!}
{
\begin{tabular}{l|cccc|cccc|cccc}
\toprule[1.25pt]
\multirow{2}{*}{\textbf{Method}} &\multicolumn{4}{c}{\textbf{MSRVTT}} &\multicolumn{4}{|c|}{\textbf{MSRVTT-\textgreater{}DiDeMo}} &\multicolumn{4}{c}{\textbf{MSRVTT-\textgreater{}LSMDC}}\\
\cmidrule(rl){2-5}\cmidrule(rl){6-9}\cmidrule(rl){10-13}
  & R@1$\uparrow$ & R@5$\uparrow$ & R@10$\uparrow$ & MdR$\downarrow$ & R@1$\uparrow$ & R@5$\uparrow$ & R@10$\uparrow$ & MdR$\downarrow$ & R@1$\uparrow$ & R@5$\uparrow$ & R@10$\uparrow$ & MdR$\downarrow$ \\ \midrule
CLIP4Clip~{\cite{luo2021clip4clip}}\ssymbol{3}~\pub{Neurocomputing22}  & {43.8} & 70.6 & 81.4 &\textbf{2.0} & 31.8 & 57.0 & 66.1 & 4.0 & 15.3 & 31.3 & 40.5 & \textbf{21.0}\\
EMCL-Net~{\cite{jin2022expectation}}\ssymbol{3}~\pub{NeurIPS22}  & {47.0} & 72.3 & 82.6 & \textbf{2.0} & 30.0 & 56.1 & 65.8 & 4.0 & 16.6 & 29.3 & 36.5 & 24.0 \\
\midrule
\rowcolor{aliceblue!60} \textbf{DiffusionRet (Ours)} & \textbf{49.0} & \textbf{75.2} & \textbf{82.7} & \textbf{2.0} & \textbf{33.2} & \textbf{59.3} & \textbf{68.4} & \textbf{3.0} & \textbf{17.1} & \textbf{32.4} & \textbf{41.0} & \textbf{21.0} \\
\bottomrule[1.25pt]
\end{tabular}}
\caption{\textbf{Text-to-video retrieval performance in out-domain retrieval settings.} ``\textit{A}-\textgreater{}\textit{B}'' indicates that ``\textit{A}'' is the source domain and ``\textit{B}'' is the target domain. ``\ssymbol{3}'' denotes our own re-implementation of baselines.}
\label{tab:out-domain}
\end{table*}

\myparagraph{Implementation Details.}
Following previous works~\cite{luo2021clip4clip,liu2022ts2,jin2022expectation,jin2022video}, we utilize the CLIP (ViT-B/32)~\cite{radford2021learning} as the pre-trained model. The dimension of the feature is 512. The temporal transformer~\cite{vaswani2017attention,li2022locality} is composed of 4-layer blocks, each including 8 heads and 512 hidden channels. The temporal position embedding and parameters are initialized from the text encoder of the CLIP. We use the Adam optimizer~\cite{kingma2014adam} and set the batch size to 128. The initial learning rate is 1e-7 for the text encoder and video encoder and 1e-3 for other modules. We set the temperature $\hat{\tau}$ to 0.01 and ${\tau}^{'}$ to 1. For short video datasets, \ie, MSRVTT, LSMDC, and MSVD, the word length is 32 and the frame length is 12. For long video datasets, \ie, ActivityNet Captions and DiDeMo, the word length is 64 and the frame length is 64. The training is divided into two stages. In the first stage, we train the feature extractor from the discrimination perspective. In the second stage, we optimize the generator from the generation perspective. For the MSRVTT and LSMDC datasets, the experiments are carried out on 2 NVIDIA Tesla V100 GPUs. For the MSVD, ActivityNet Captions, and DiDeMo datasets, the experiments are carried out on 8 NVIDIA Tesla V100 GPUs. In both of the tasks of text-to-video and video-to-text retrieval, we assume that only the candidate sets are known in advance. In the inference phase, we consider both the distance of video and text representations in the representation space and the joint probability of video and text.

\subsection{Comparison with State-of-the-art}
We compare the proposed DiffusionRet with other methods on five benchmarks. In Tab.~{\ref{MSRVTT results}}, we show the retrieval results on the MSRVTT dataset. Our model outperforms the recently proposed state-of-the-art methods on both text-to-video retrieval and video-to-text retrieval tasks. Tab.~{\ref{Experiments0}} shows text-to-video retrieval results on the LSMDC, MSVD, ActivityNet Captions, and DiDeMo datasets. DiffusionRet achieves consistent improvements across different datasets, which demonstrates the effectiveness of our method.

\subsection{Ablation Study}
\myparagraph{Generation loss type.} In Tab.~\ref{tab:loss}, we compare two common generation losses, \ie, mean-squared loss~(MSE) and Kullback-Leibler~(KL) divergence. Results demonstrate that the KL divergence achieves optimal retrieval performance. We explain that it is because KL divergence can better measure the distance between probabilities than MSE, so it is more suitable for probability generation.

\myparagraph{Sampling strategy.} Denoising diffusion probabilistic models~\cite{ho2020denoising}~(DDPM) learn the underlying data distribution by a Markov chain as shown in Eq.~\ref{Markov}. To accelerate the sampling process of diffusion models, denoising diffusion implicit models~\cite{song2020denoising}~(DDIM) formulate a Markov chain that reverses a non-Markovian perturbation process. As shown in Tab.~\ref{tab:sampling}, we find that the two sampling strategies have similar performance. To speed up the sampling process, we use DDIM by default in practice.

\myparagraph{Schedule of $\beta$.} The schedule of $\beta$ controls how the step size increases. We compare the linear schedule and cosine schedule in Tab.~\ref{tab:schedule}. As shown in Tab.~\ref{tab:schedule}, we find that the cosine schedule performs well, so we adopt the cosine schedule by default, which is the same as the default setting in the motion generation task~\cite{tevet2022human}.

\myparagraph{Training strategy.} In Tab.~\ref{tab:training}, we compare different training strategies. Similar to the findings of the pioneer work~\cite{bernardo2007generative}, we find that pure discriminant training has better performance on limited data than pure generative training. The hybrid training method we proposed achieves the best results, which indicates that our hybrid training strategy can combine the advantages of the two training methods.

\myparagraph{The number of steps.} In Tab.~\ref{tab:step}, we explore the influence of the number of diffusion steps. Results demonstrate that the number of steps of 50 achieves optimal performance in the retrieval task, outperforming the standard value of 1000 of the image generation task~\cite{ho2020denoising,dhariwal2021diffusion}. We consider that this is due to the simpler probability distribution in retrieval compared to the pixel distribution in natural images. Therefore, compared with the image generation task, the cross-modal retrieval task only needs fewer diffusion steps.

\myparagraph{Scale of $\beta$.} The scale of $\beta$ indicates the signal-to-noise ratio of the diffusion process~\cite{chen2022diffusiondet}. We evaluate the scale range $[0.1, 2.0]$ as shown in Tab.~\ref{tab:scale}. We find that the model achieves the best performance at the scaling of 1.0, so we set the scale of $\beta$ to 1.0 in practice, which is the same as the default setting in the image generation task~\cite{dhariwal2021diffusion}.

\subsection{The Efficiency of DiffusionRet}
We provide comparisons of the inference time and memory consumption of our method with other methods under different conditions (the number of diffusion steps and the size of the candidate gallery) in Tab.~\ref{tab:cost}. As shown in Tab.~\ref{tab:cost}, our method is as efficient as the existing state-of-the-art methods during the inference stage, which we explain in the following three aspects. \textbf{(1) Lightweight denoising network.} Our denoising network is lightweight (2.50 M parameter), while other methods use complex matching networks, \eg, TS2-Net which uses the token shift transformer and token selection transformer for fine-grained alignment. \textbf{(2) Efficient feature extractor.} About 80\% of the inference time is spent on feature extraction such as extracting query text features. Compared with other methods with bulky feature extractors, \eg, X-Pool which uses an additional transformer-based pooling module to aggregate features, our method only uses vanilla transformer to extract features and is therefore more efficient. \textbf{(3) Scalability.} We can increase the number of diffusion steps to boost performance at a negligible time cost, indicating the scalability of our method. Compared with other methods, our method is more flexible and better suited to different retrieval scenarios where accuracy and speed are required differently.

\begin{figure*}[tbp]
\centering
\includegraphics[width=1.\linewidth]{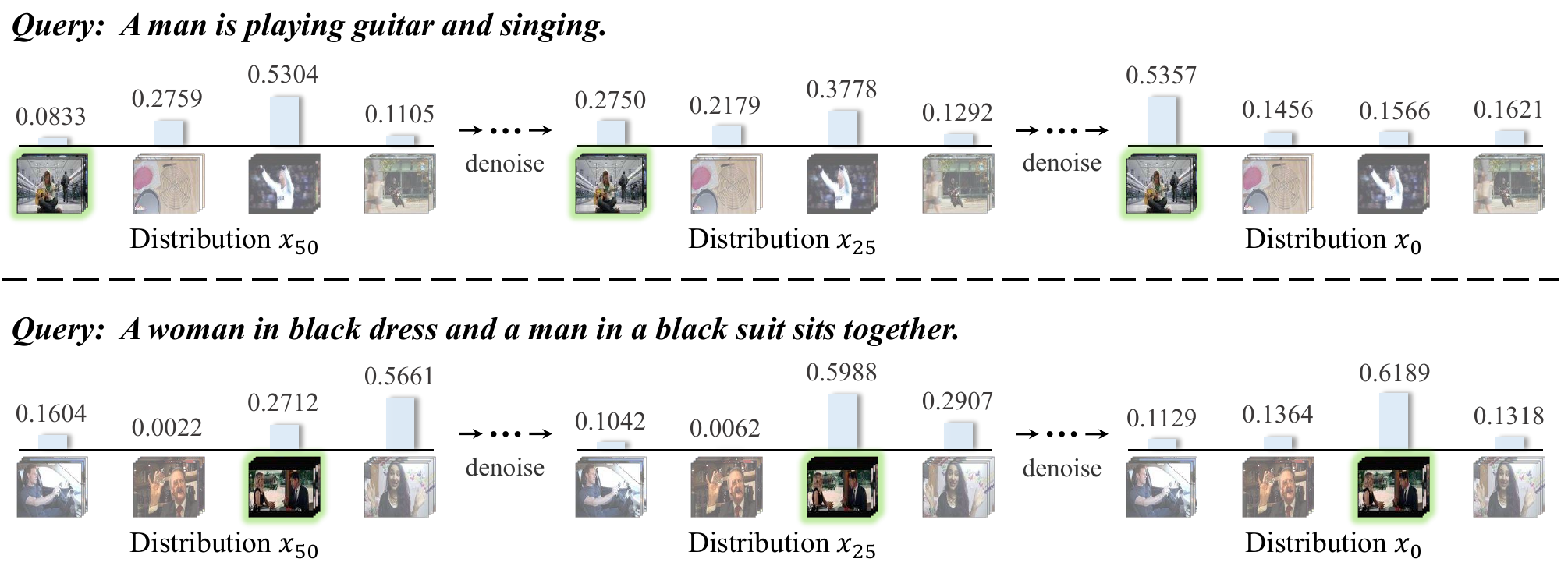}
\vspace{-1.5em}
\caption{\textbf{The visualization of the diffusion process of the probability distribution.} We highlight the ground truth in green, and show the process from randomly initialized noise input ($x_{50}$) to the final predicted distribution ($x_0$).}
\label{fig:v}
\end{figure*}

\subsection{Out-domain Retrieval}
Current text-video retrieval methods are mainly evaluated on the same dataset. To further evaluate the generalization to unseen data, we conduct out-domain retrieval~\cite{chen2020fine}: we first pre-train a model on one dataset (the ``source'') and then measure its performance on another dataset (the ``target'') that is unseen in the training. As shown in Tab.~\ref{tab:out-domain}, we compare the proposed DiffusionRet with other baselines (\ie, CLIP4Clip~\cite{luo2021clip4clip} and EMCL-Net~\cite{jin2022expectation}) in out-domain retrieval settings. We find that the discriminant approaches fail to migrate the performance of in-domain retrieval to out-of-domain retrieval well. For example, the performance of EMCL-Net is significantly higher than that of CLIP4Clip in in-domain retrieval. However, in out-domain retrieval, the performance of EMCL-Net is slightly lower than that of CLIP4Clip. In contrast, DiffusionRet performs well for both in-domain and out-of-domain retrieval.

To further illustrate the benefits of our generative modeling, we provide the visualization of the similarity distribution in in-domain retrieval and out-domain retrieval. As shown in Fig.~\ref{fig:p}, compared to the baseline models, our DiffusionRet maximally keeps the distributions of positive and negative pairs separate from each other in both in-domain and out-domain settings. These results confirm that our method is more generalizable and transferrable for the unseen data than typical discriminant models.

\subsection{Qualitative Analysis}
The coarse-to-fine nature of the diffusion model enables it to progressively uncover the correlation between text and video, rendering it an effective approach for cross-modal retrieval. To better understand the diffusion process, we show the additional visualization of the diffusion process in Fig.~\ref{fig:v}. These results demonstrate that our method can progressively uncover the correlation between text and video.

\subsection{Why Diffusion Models}
Diffusion models have demonstrated remarkable generative power in various fields. Besides the powerful generative power of diffusion models, we explain other advantages of applying the diffusion model rather than other generative approaches to cross-modal retrieval, mainly in two aspects. \textbf{First}, the coarse-to-fine nature of the diffusion model enables it to progressively uncover the correlation between text and video, rendering it a more effective approach for retrieval tasks than other generation training methods, such as generative adversarial network~\cite{goodfellow2020generative} and variational autoencoder~\cite{kingma2013auto}. \textbf{Second}, the many-to-many nature of the diffusion model makes it more suitable for generating joint probabilities than the auto-regressive networks~\cite{frey1995does,radford2018improving}. In Fig.~\ref{fig:v}, we show the visualization of the diffusion process. These results further demonstrate the above two advantages of applying the diffusion model rather than other generative approaches to cross-modal retrieval.

\section{Conclusion}
In this paper, we propose DiffusionRet, the first generative diffusion-based framework for text-video retrieval. By explicitly modeling the joint probability distribution of text and video, DiffusionRet shows promise to solve the intrinsic limitations of the current discriminant regime. It successfully optimizes the DiffusionRet from both the generation perspective and discrimination perspective. This makes DiffusionRet principled and well-applicable in both in-domain retrieval and out-domain retrieval settings. To the best of our knowledge, we are the first to tackle text-video retrieval from the generative viewpoint. Besides, we show that our generation modeling method is superior to existing discriminant methods in terms of performance and generalization ability. This work is the first effort to promote generative methods for text-video retrieval, which may be meaningful and helpful to the community.

\noindent \textbf{Acknowledgements.} This work was supported in part by the National Key R\&D Program of China (No. 2022ZD0118201), Natural Science Foundation of China (No. 61972217, 32071459, 62176249, 62006133, 62271465, 62202014). Li Yuan is also sponsored by CCF Tencent Open Research Fund.

{\small
\bibliographystyle{ieee_fullname}
\bibliography{ref}
}

\appendix

\renewcommand{\thefootnote}{\fnsymbol{footnote}}

\renewcommand{\thetable}{\Alph{table}}
\renewcommand{\theequation}{\Alph{equation}}
\renewcommand{\thefigure}{\Alph{figure}}

\setcounter{table}{0}
\setcounter{section}{0}
\setcounter{figure}{0}
\setcounter{equation}{0}

\begin{table*}[htb]
\centering
\footnotesize
\resizebox{1.\linewidth}{!}
{
\begin{tabular}{l|cccccc|cccccc}
\toprule[1.5pt]
\multirow{2}{*}{\textbf{Method}} &  \multicolumn{6}{c|}{\textbf{LSMDC}}        & \multicolumn{6}{c}{\textbf{MSVD}}  \\ 
\cmidrule(rl){2-7}\cmidrule(rl){8-13}
  & R@1$\uparrow$ & R@5$\uparrow$ & R@10$\uparrow$ & Rsum$\uparrow$ & MdR$\downarrow$ & MnR$\downarrow$
& R@1$\uparrow$ & R@5$\uparrow$ & R@10$\uparrow$ & Rsum$\uparrow$ & MdR$\downarrow$ & MnR$\downarrow$    \\ \midrule
TT-CE~{\cite{croitoru2021teachtext}}~\pub{ICCV21}  & 17.5 & 36.0 & 45.0 & 98.5 & 14.3 & - & 27.1 & 55.3 & 67.1 & 149.5 & 4.0 & - \\
CLIP4Clip~{\cite{luo2021clip4clip}}~\pub{Neurocomputing22}  & 20.8 & 39.0 & 48.6 & 108.4 & 12.0 & 54.2 & 62.0 & 87.3 & 92.6 & 241.9 & \Frst{1.0} & \textbf{4.3} \\
EMCL-Net~{\cite{jin2022expectation}}~\pub{NeurIPS22}  & 22.2 & 40.6 & 49.2 & 112.0 & 12.0 & - & 54.3 & 81.3 & 88.1 & 223.7 & \textbf{1.0} & 5.6 \\
\midrule
\rowcolor{aliceblue!60} \textbf{DiffusionRet (Ours)}  & \Frst{23.0} & \Frst{43.5} & {51.5} & \Frst{118.0} & \Frst{9.0} & {40.2} & \textbf{61.9} & \Frst{88.3} & \Frst{92.9} & \Frst{243.1} & \textbf{1.0} & 4.5     \\
\rowcolor{aliceblue!60} + QB-Norm~\cite{bogolin2022cross} & {22.8} & {43.2} & \textbf{51.6} & {117.6} &\textbf{9.0} & \textbf{40.0} & {60.3} & {86.4} & {92.0} & {238.7} & \textbf{1.0} & 4.5 \\
\midrule[1.25pt]
\multirow{2}{*}{\textbf{Method}} &  \multicolumn{6}{c|}{\textbf{ActivityNet Captions}}        & \multicolumn{6}{c}{\textbf{DiDeMo}}  \\ 
\cmidrule(rl){2-7}\cmidrule(rl){8-13}
  & R@1$\uparrow$ & R@5$\uparrow$ & R@10$\uparrow$ & Rsum$\uparrow$ & MdR$\downarrow$ & MnR$\downarrow$
& R@1$\uparrow$ & R@5$\uparrow$ & R@10$\uparrow$ & Rsum$\uparrow$ & MdR$\downarrow$ & MnR$\downarrow$    \\ \midrule
TT-CE~{\cite{croitoru2021teachtext}}~\pub{ICCV21}  & 23.0 & 56.1 & - & - & 4.0 & - & 21.1 & 47.3 & 61.1 & 129.5 & 6.3 & - \\
CLIP4Clip~{\cite{luo2021clip4clip}}~\pub{Neurocomputing22}  & 41.4 & 73.7 & 85.3 & 200.4 & \textbf{2.0} & 6.7 & 41.4 & 68.2 & 79.1 & 188.7 & {2.0} & 12.4 \\
EMCL-Net~{\cite{jin2022expectation}}~\pub{NeurIPS22}  & 42.7 & 74.0 & - & - & \textbf{2.0} & - & 45.7 & 74.3 & 82.7 & 202.7 & 2.0 & 10.9 \\
HBI~\cite{jin2022video}~\pub{CVPR23} & 42.4 & 73.0 & 86.0 & 201.4 & \textbf{2.0} & 6.5 & 46.2 & 73.0 & 82.7 & 201.9 & 2.0 & \textbf{8.7} \\
\midrule
\rowcolor{aliceblue!60} \textbf{DiffusionRet (Ours)}  & {43.8} & {75.3} & \Frst{86.7} & {205.8} & \Frst{2.0} & \Frst{6.3} & 46.2 & {74.3} & {82.2} & 202.7 & {2.0} & 10.7      \\
\rowcolor{aliceblue!60} + QB-Norm~\cite{bogolin2022cross} & \textbf{47.4} & \textbf{76.3} & \textbf{86.7} & \textbf{210.4} &\textbf{2.0} & {6.7} & \textbf{50.3} &\textbf{75.1} &\textbf{82.9} & \textbf{208.3} &\textbf{1.0} & 10.3 \\ \bottomrule[1.5pt]
\end{tabular}
}
\caption{\textbf{Video-to-text retrieval performance on the LSMDC, MSVD, and ActivityNet Captions datasets.} ``$\uparrow$'' denotes that higher is better. ``$\downarrow$'' denotes that lower is better.}
\label{tab:video-to-text}
\end{table*}

\begin{table*}[t]
\footnotesize
\centering
\resizebox{1.\linewidth}{!}
{
\begin{tabular}{l|ccccc|ccccc}
\toprule[1.25pt]
\multirow{2}{*}{\textbf{Method}}  &\multicolumn{5}{c}{\textbf{MSRVTT}} &\multicolumn{5}{|c}{\textbf{MSRVTT-\textgreater{}ActivityNet Captions}}\\
\cmidrule(rl){2-6}\cmidrule(rl){7-11}
  & R@1$\uparrow$ & R@5$\uparrow$ & R@10$\uparrow$ & Rsum$\uparrow$ & MdR$\downarrow$ 
& R@1$\uparrow$ & R@5$\uparrow$ & R@10$\uparrow$ & Rsum$\uparrow$ & MdR$\downarrow$  \\ \midrule
CLIP4Clip~{\cite{luo2021clip4clip}}\ssymbol{3}~\pub{Neurocomputing22}  & 43.8 & 70.6 & 81.4 & 195.8 & \textbf{2.0} & {29.1} & 58.3 & 72.1 & 159.5 & 4.0  \\
EMCL-Net~{\cite{jin2022expectation}}\ssymbol{3}~\pub{NeurIPS22} & 47.0 & 72.3 & 82.6 & 201.9 & \textbf{2.0} & {28.7} & 56.8 & 70.6 & 156.1 & 4.0\\
\midrule
\rowcolor{aliceblue!60} \textbf{DiffusionRet (Ours)} & \textbf{49.0} & \textbf{75.2} & \textbf{82.7} & \textbf{206.9} & \textbf{2.0}  & \textbf{31.5} & \textbf{60.0} & \textbf{73.8} & \textbf{165.3} & \textbf{3.0} \\
\bottomrule[1.25pt]
\end{tabular}
}
\caption{\textbf{Text-to-video retrieval performance in out-domain retrieval settings.} ``\textit{MSRVTT}-\textgreater{}\textit{ActivityNet Captions}'' denotes that the generalization results on unseen ActivityNet Captions test setting using pre-trained models on the MSRVTT dataset.``\ssymbol{3}'' denotes our own re-implementation of baselines. ``$\uparrow$'' denotes that higher is better. ``$\downarrow$'' denotes that lower is better.}
\label{tab:out-domain_others}
\end{table*}

\section{Appendix}
This appendix provides the descriptions of datasets (Sec.~\ref{apendix:dataset}), implementation details (Sec.~\ref{apendix:Implementation}), video-to-text retrieval performance on the LSMDC, MSVD, and ActivityNet Captions datasets (Sec.~\ref{appendix:video-to-text}), additional experiments for the out-domain retrieval (Sec.~\ref{appendix:out-domain}), the reason for applying diffusion models (Sec.~\ref{why_diffusion}), discussion of the limitations (Sec.~\ref{appendix:limitations}), the additional visualization of the diffusion process (Sec.~\ref{appendix:diffusion_process}), the visualization of the text-frame attention map (Sec.~\ref{appendix:t_f_attention}), and the visualization of the text-to-video retrieval examples (Sec.~\ref{appendix:text-to-video}).

\subsection{Datasets and Implementation Details}
\subsubsection{Datasets}\label{apendix:dataset}
We compare the proposed DiffusionRet with other methods on five benchmark text-video retrieval datasets, including MSRVTT~\cite{xu2016msr}, LSMDC~\cite{rohrbach2015a}, MSVD~\cite{chen2011collecting}, ActivityNet Captions~\cite{krishna2017dense}, and DiDeMo~\cite{anne2017localizing}. 

\myparagraph{MSRVTT.} MSRVTT~\cite{xu2016msr} contains 10,000 YouTube videos, each with 20 text descriptions. We follow the training protocol in \cite{liu2019use,gabeur2020multi,miech2019howto100m} and evaluate on text-to-video and video-to-text search tasks on the 1K-A testing split with 1K video or text candidates defined by \cite{yu2018a}.

\myparagraph{LSMDC.} LSMDC~\cite{rohrbach2015a} contains 118,081 video clips from 202 movies. The duration of videos in the LSMDC dataset is short. We follow the split of \cite{gabeur2020multi} with 1,000 videos for testing. 

\myparagraph{MSVD.} MSVD~\cite{chen2011collecting} contains 1,970 videos. Each video has approximately 40 associated text description. Videos in the MSVD dataset are short in duration, lasting about 10 to 25 seconds. We follow the official split of 1,200 and 670 as the train and test set, respectively. 

\myparagraph{ActivityNet Captions.} ActivityNet Captions~\cite{krishna2017dense} consists densely annotated temporal segments of 20K YouTube videos. Following~\cite{gabeur2020multi,patrick2021support,wang2021t2vlad}, we concatenate descriptions of segments in a video to construct ``video-paragraph'' for retrieval. We report results on the ``val1'' split of 10,009 and 4,917 as the train and test set. 

\myparagraph{DiDeMo.} DiDeMo~\cite{anne2017localizing} contains 10,464 videos annotated 40,543 text descriptions. We concatenate descriptions of segments in a video to construct ``video-paragraph'' for retrieval. We follow the training and evaluation protocol in~\cite{luo2021clip4clip}.

\begin{figure*}[tbp]
\centering
\includegraphics[width=1.\linewidth]{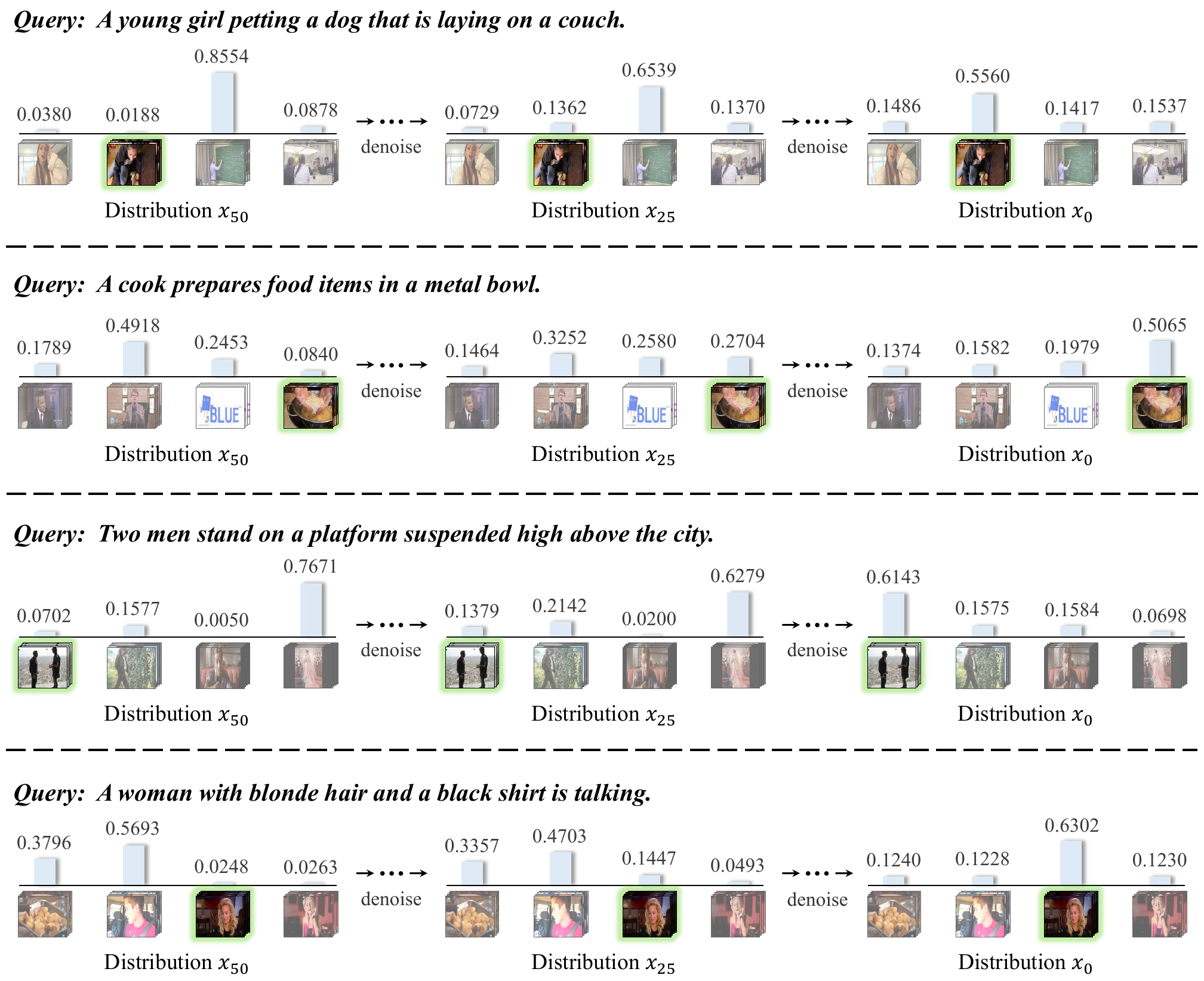}
\caption{\textbf{The visualization of the diffusion process of the probability distribution.} We highlight the ground truth in green, and show the process from randomly initialized noise input ($x_{50}$) to the final predicted distribution ($x_0$). The iterative refinement property and many-to-many nature of the diffusion model render it an effective approach for text-video retrieval.}
\label{fig:v_more}
\end{figure*}

\begin{figure*}[tbp]
\centering
\includegraphics[width=1.\linewidth]{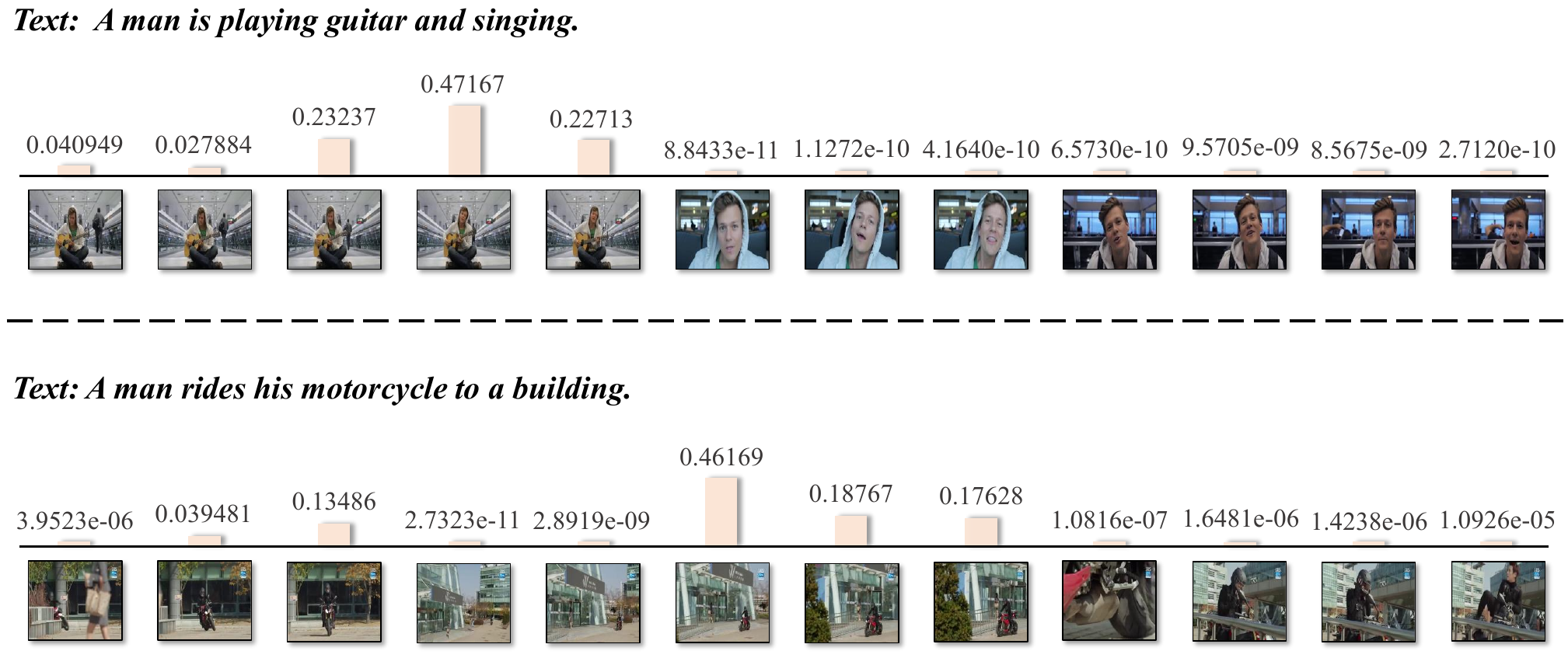}
\caption{\textbf{The visualization of the text-frame attention map.} These results demonstrate that our method can capture the correlation between text and frames.}
\label{fig:t_f_attention}
\end{figure*}

\begin{figure*}[tbp]
\centering
\includegraphics[width=1.\linewidth]{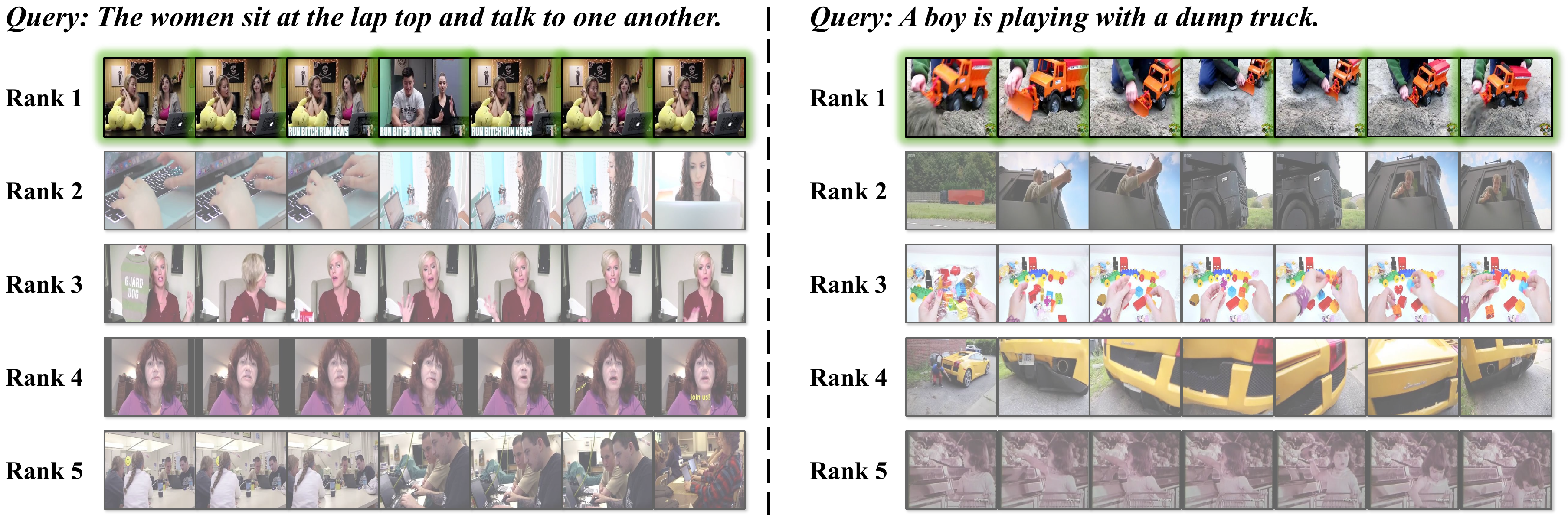}
\caption{\textbf{The visualization of the text-to-video results.} We highlight the ground truth in green. These results demonstrate that our method can mine the correlation between text and video effectively.}
\label{fig:retrieval}
\end{figure*}

\subsubsection{Implementation Details.}\label{apendix:Implementation}
Following previous works~\cite{luo2021clip4clip,jin2022expectation,jin2022video,jin2023text}, we utilize the CLIP (ViT-B/32)~\cite{radford2021learning} as the pre-trained model. The dimension of the feature is 512. The temporal transformer~\cite{vaswani2017attention,li2022locality} is composed of 4-layer blocks, each including 8 heads and 512 hidden channels. The temporal position embedding~\cite{yu2022position} and parameters are initialized from the text encoder of the CLIP. We use the Adam optimizer~\cite{kingma2014adam} and set the batch size to 128. The initial learning rate is 1e-7 for the text encoder and video encoder and 1e-3 for other modules. We set the temperature $\hat{\tau}$ to 0.01 and ${\tau}^{'}$ to 1. For short video datasets, \ie, MSRVTT, LSMDC, and MSVD, the word length is 32 and the frame length is 12. For long video datasets, \ie, ActivityNet Captions and DiDeMo, the word length is 64 and the frame length is 64. 

The training is divided into two stages. In the first stage, we train the feature extractor from the discrimination perspective. In the second stage, we optimize the generator from the generation perspective. For the MSRVTT and LSMDC datasets, the experiments are carried out on 2 NVIDIA Tesla V100 GPUs. For the MSVD, ActivityNet Captions, and DiDeMo datasets, the experiments are carried out on 8 NVIDIA Tesla V100 GPUs. In both of the tasks of text-to-video and video-to-text retrieval, we assume that only the candidate sets are known in advance. In the inference phase, we consider both the distance of video and text representations in the representation space and the joint probability of video and text. Code is available at \href{https://github.com/jpthu17/DiffusionRet}{https://github.com/jpthu17/DiffusionRet}.

\subsection{Additional Results and Discussions}
\subsubsection{Video-to-Text Retrieval}\label{appendix:video-to-text}
We compare the proposed DiffusionRet with other meth-
ods on five benchmark. In addition to the text-to-video retrieval results in the main paper, we provide video-to-text retrieval results on the LSMDC, MSVD, ActivityNet Captions, and DiDeMo datasets in Tab.~{\ref{tab:video-to-text}}. Extensive experiments on five datasets, including MSRVTT, LSMDC, MSVD, ActivityNet Captions, and DiDeMo, demonstrate that our method is capable of dealing with both short and long videos. DiffusionRet achieves consistent improvements across different datasets, which demonstrates the effectiveness of our method. 

\subsubsection{Out-domain Retrieval}\label{appendix:out-domain}
Most text-video retrieval methods~\cite{luo2021clip4clip,jin2022expectation,jin2022video,jin2023text} are evaluated using the same dataset, which may not reflect their ability to generalize to unseen data. To this end, we perform out-domain retrieval by pre-training a model on one dataset (referred to as the ``source'') and evaluating its performance on another dataset (referred to as the ``target'') that is not included in the training. In addition to the out-domain retrieval experiments in the main paper, we provide additional experiments in the out-domain retrieval setting (MSRVTT-\textgreater{}ActivityNet Captions) in Tab.~\ref{tab:out-domain_others}. We find that discriminant approaches do not transfer well from in-domain to out-of-domain retrieval. For instance, EMCL-Net outperforms CLIP4Clip in in-domain retrieval, but its performance is slightly lower than CLIP4Clip in out-domain retrieval. In contrast, DiffusionRet achieves good performance in both in-domain and out-of-domain retrieval.

\subsubsection{Why Diffusion Models}\label{why_diffusion}
Diffusion models have demonstrated remarkable generative power in various fields. Besides the powerful generative power of diffusion models, we explain other advantages of applying the diffusion model rather than other generative approaches to cross-modal retrieval, mainly in two aspects. \textbf{First}, the coarse-to-fine nature of the diffusion model enables it to progressively uncover the correlation between text and video, rendering it a more effective approach for retrieval tasks than other generation training methods, such as generative adversarial network~\cite{goodfellow2020generative} and variational autoencoder~\cite{kingma2013auto}. \textbf{Second}, the many-to-many nature of the diffusion model makes it more suitable for generating joint probabilities than the auto-regressive networks~\cite{frey1995does,radford2018improving}. We recommend further investigation of the potential of the generative method for discriminant tasks in future research. In our future work, we will explore our algorithm in segmentation~\cite{li2022dynamic,li2023multi} and visual question answering~\cite{li2022joint,li2023weakly,li2023tg}.

\subsubsection{Limitations of our Work}\label{appendix:limitations}
Generative models have focused on generative tasks, \eg, image generation~\cite{ho2020denoising,song2020denoising}, natural language generation~\cite{austin2021structured,li2022diffusion}, and audio generation~\cite{popov2021grad}. Some other works have attempted to adapt the generative models for discriminant tasks, \eg, image segmentation~\cite{amit2021segdiff}, visual grounding~\cite{cheng2023parallel}, and detection~\cite{chen2022diffusiondet}. However, these precursor methods require additional discriminative training. To train on limited data, we optimize the proposed generation model from both generation and discrimination perspectives. Although such a hybrid training method can improve model performance with limited data, we believe that pure generative training is a more promising solution when the data is sufficient. We suggest exploring a pure generative training approach to the retrieval problem in the future.

\subsection{Additional Visualizations}
\subsubsection{Diffusion Process}\label{appendix:diffusion_process}
The coarse-to-fine nature of the diffusion model enables it to progressively uncover the correlation between text and video, rendering it an effective approach for cross-modal retrieval. To better understand the diffusion process, we show the additional visualization of the diffusion process in Fig.~\ref{fig:v_more}. These results demonstrate that our method can progressively uncover the correlation between text and video.

\subsubsection{Text-Frame Attention Map}\label{appendix:t_f_attention}
To extract the joint encoding of text and video, we propose the text-frame attention encoder, which takes text representation as query and frame representation as key and value. To better understand the process of joint encoding of text and video, we show the visualization of the text-frame attention map in Fig.~\ref{fig:t_f_attention}. As shown in Fig.~\ref{fig:t_f_attention}, the text-frame attention encoder adaptively extracts the frames that are similar to the text so that fine-grained video features can be extracted. These results demonstrate that our method can capture the correlation between text and frames.

\subsubsection{Text-to-Video Retrieval}\label{appendix:text-to-video}
We show two retrieval examples from the MSRVTT testing set for text-to-video retrieval in Fig.~\ref{fig:retrieval}. As shown in Fig.~\ref{fig:retrieval}, our method successfully retrieves the ground-truth video. These results demonstrate that our method can mine the correlation between text and video effectively.

\end{document}